\documentclass[runningheads]{llncs}

\usepackage{eccv}

\usepackage{eccvabbrv}

\usepackage{graphicx}
\usepackage{booktabs}

\usepackage{graphicx}
\usepackage{amsmath}
\usepackage{amssymb}
\usepackage{arydshln}
\usepackage{booktabs}
\usepackage{xcolor}
\usepackage{soul}

\newcommand{\hlfancy}[2]{\sethlcolor{#1}\hl{#2}}

\colorlet{soulred}{red!20}
\colorlet{soulgreen}{green!20}
\colorlet{soulblue}{blue!20}
\colorlet{soulorange}{orange!20}

\usepackage[accsupp]{axessibility}  

\usepackage[pagebackref,breaklinks,colorlinks,citecolor=eccvblue]{hyperref}

\usepackage{orcidlink}
\usepackage{xargs}
\usepackage[colorinlistoftodos,textsize=tiny]{todonotes}
\newcommandx{\todoc}[2][1=]{{\todo[linecolor=orange,backgroundcolor=orange!25,bordercolor=orange,#1]{\tiny
      TODO: #2}}}
\newcommandx{\unsure}[2][1=]{{\todo[linecolor=red,backgroundcolor=red!25,bordercolor=red,#1]{\tiny
      UNSURE: #2}}}
\newcommandx{\change}[2][1=]{{\todo[linecolor=blue,backgroundcolor=blue!25,bordercolor=blue,#1]{\tiny
      CHANGE: #2}}}
\newcommandx{\info}[2][1=]{{\todo[linecolor=green,backgroundcolor=green!25,bordercolor=green,#1]{\tiny
      INFO: #2}}}
\newcommandx{\improvement}[2][1=]{{\todo[linecolor=violet,backgroundcolor=violet!25,bordercolor=violet,#1]{\tiny
      IMPROVEMENT: #2}}}
\newcommandx{\thiswillnotshow}[2][1=]{{\todo[disable,#1]{THIS WILL NOT SHOW:
      #2}}}
      
\begin{document}

\title{Representation Learning in a Decomposed Encoder Design for Bio-inspired Hebbian Learning } 

\titlerunning{Decomposed Encoders for Hebbian Learning}

\author{
    Achref Jaziri\inst{1} \and
    Sina Ditzel\inst{1} \and
    Iuliia Pliushch\inst{2} \and
    Visvanathan Ramesh\inst{1}\orcidlink{0000-0002-8842-905X}
}

\authorrunning{A. Jaziri et al.}

\institute{
    Department of Computer Science and Mathematics, Goethe University, 60323 Frankfurt am Main, Germany \and
    Department of Educational Psychology, Goethe University, 60629 Frankfurt am Main, Germany \\
    \email{Jaziri@em.uni-frankfurt.de, sina.ditzel@flandersmake.be, pliushch@psych.uni-frankurt.de, vramesh@em.uni-frankfurt.de}
}


\maketitle

\begin{abstract}

Modern data-driven machine learning system designs exploit inductive biases in architectural structure, invariance and equivariance requirements, task-specific loss functions, and computational optimization tools. Previous works have illustrated that human-specified quasi-invariant filters can serve as a powerful inductive bias in the early layers of the encoder, enhancing robustness and transparency in learned classifiers. This paper explores this further within the context of representation learning with bio-inspired Hebbian learning rules. We propose a modular framework trained with a bio-inspired variant of contrastive predictive coding, comprising parallel encoders that leverage different invariant visual descriptors as inductive biases. We evaluate the representation learning capacity of our system in classification scenarios using diverse image datasets (GTSRB, STL10, CODEBRIM) and video datasets (UCF101). Our findings indicate that this form of inductive bias significantly improves the robustness of learned representations and narrows the performance gap between models using local Hebbian plasticity rules and those using backpropagation, while also achieving superior performance compared to non-decomposed encoders.
\end{abstract}

\section{Introduction}

Learning in the primate brain is believed to occur in an unsupervised manner, governed by local plasticity rules \cite{hebb1949organization}. Recent research aims to incorporate these brain-inspired learning rules into deep neural networks (DNNs) as an alternative to backpropagation \cite{pogodin2020kernelized,lillicrap2016random,oja1982simplified}. The main motivation is to understand brain learning mechanisms and develop more computationally efficient systems \cite{wunderlich2019demonstrating}. However, DNNs trained with current bio-inspired rules often achieve worse performance compared to those trained using standard backpropagation \cite{bartunov2018assessing}.

We propose that integrating local plasticity rules within a modular, decomposable structure can help narrow the performance gap with backpropagation-trained networks. Drawing on the idea that biological learning doesn't start from a blank slate \cite{zador2019critique}, we explore the synergy between inductive biases, such as visual invariant operators from physics and mathematical analyses, and biologically inspired local learning rules.

 DNN implicitly learn transformations from data to solve a certain classification rather than explicitly defining invariances \cite{bengio2009learning,krizhevsky2012imagenet,lecun2004learning}. While DNNs excel in many applications, their black-box nature makes it difficult to identify the learned invariances and features used for decision-making. This lack of transparency, along with vulnerability to adversarial perturbations, presents significant challenges for the safe deployment of systems based on DNNs \cite{BarredoArrieta2020,Carvalho2019}.

To better understand and mitigate these issues, some works have suggested adopting a decomposable design \cite{Lipton2016}. Computer vision tasks often require invariance to variables such as object pose, size, and illumination. In the late 1980s and 1990s, expert-driven model-based designs achieved this by stacking and combining quasi-invariant transformations, ensuring the output remained stable despite irrelevant changes in context \cite{binford1993quasi,chin1986model}.

These efforts highlight the importance of selecting appropriate invariant operators to disregard nuisance variables. This concept, combining model-based and deep learning approaches, leverages task-specific inductive biases while learning parts that are difficult to model \cite{Baslamisli2021}. For instance, \cite{Luan2018,Perez2020} showed that incorporating human-specified quasi-invariant filters, such as Gabor filters, in the early layers of neural networks can enhance robustness and transparency.

Parallels exist between decomposable vision systems and brain-inspired architectures. The mammalian visual system processes different input signals through parallel, hierarchical pathways \cite{ungerleider1982two}. Brain-inspired designs mimic these structures, using specialized pathways to process multiple cues like color, shape, and texture in parallel \cite{von2014vision,hinton2021represent,hawkins2019framework}.

We present a framework that incorporates multiple encoder networks, each augmented with distinct invariant visual descriptors such as red-green normalization, local binary pattern operator, and dual-tree complex wavelet transform. These networks are trained using a contrastive loss in a self-supervised manner. To validate the learned representations, we employ a linear classifier on various image datasets (GTSRB, STL10, CODEBRIM) and video datasets (UCF101). For video data, an additional encoder is trained specifically to learn motion-sensitive representations. The video data experiments are in the appendix.

Our findings demonstrate that the careful selection of inductive biases for decomposition significantly enhances the classification of learned representations. This approach is particularly effective with bio-inspired learning rules, helping to bridge the performance gap with backpropagation methods while also achieving better overall performance compared to non-decomposed encoders\footnote{Code is accessible here: \url{https://github.com/achrefjaziri/DecomposedEncoders}}.

\section{Related Work}

\textbf{DNNs with Invariant Operators:} Various works aim to create more transparent and robust pipelines by integrating well-understood operators into data-driven systems \cite{oyallon2017scaling,li2020wavelet}. For example, the SIFT detector, quasi-invariant to scale, orientation, and illumination, has been combined with neural architectures \cite{perronnin2015fisher}. Based on the Local Binary Pattern (LBP) operator, an adapted convolutional layer with fewer parameters was proposed to extract texture features \cite{juefei2017local,Lin2020}. Evidence shows that simulating the primate primary visual cortex (V1) in early CNN layers increases robustness against input perturbations and adversarial attacks \cite{dapello2020simulating}. 


\textbf{Hebbian Learning:} Unlike backpropagation, Hebbian plasticity rules are local, depending only on the activations of pre- and post-synaptic neurons, and often a third factor related to reward or other high-level signals. Recent research leverages these local plasticity rules to train deep neural networks, offering alternatives that decouple feedforward and feedback paths to address the biological implausibility of backpropagation \cite{bengio2009learning,krotov2019unsupervised,nokland2019training,pogodin2020kernelized,xie2003equivalence,millidge2022backpropagation,amit2019deep,lillicrap2016random,han2019efficient,moskovitz2018feedback}.

While these methods often result in decreased performance on downstream tasks, scaling beyond simple image classification remains a challenge for bio-inspired systems \cite{bartunov2018assessing}. In this work, we show that the performance gap of bio-inspired learning rules can be mitigated with appropriate network structures and inductive biases.



\section{Method}

\begin{figure*}[htbp]
	\centering
	\includegraphics[width=0.8\linewidth]{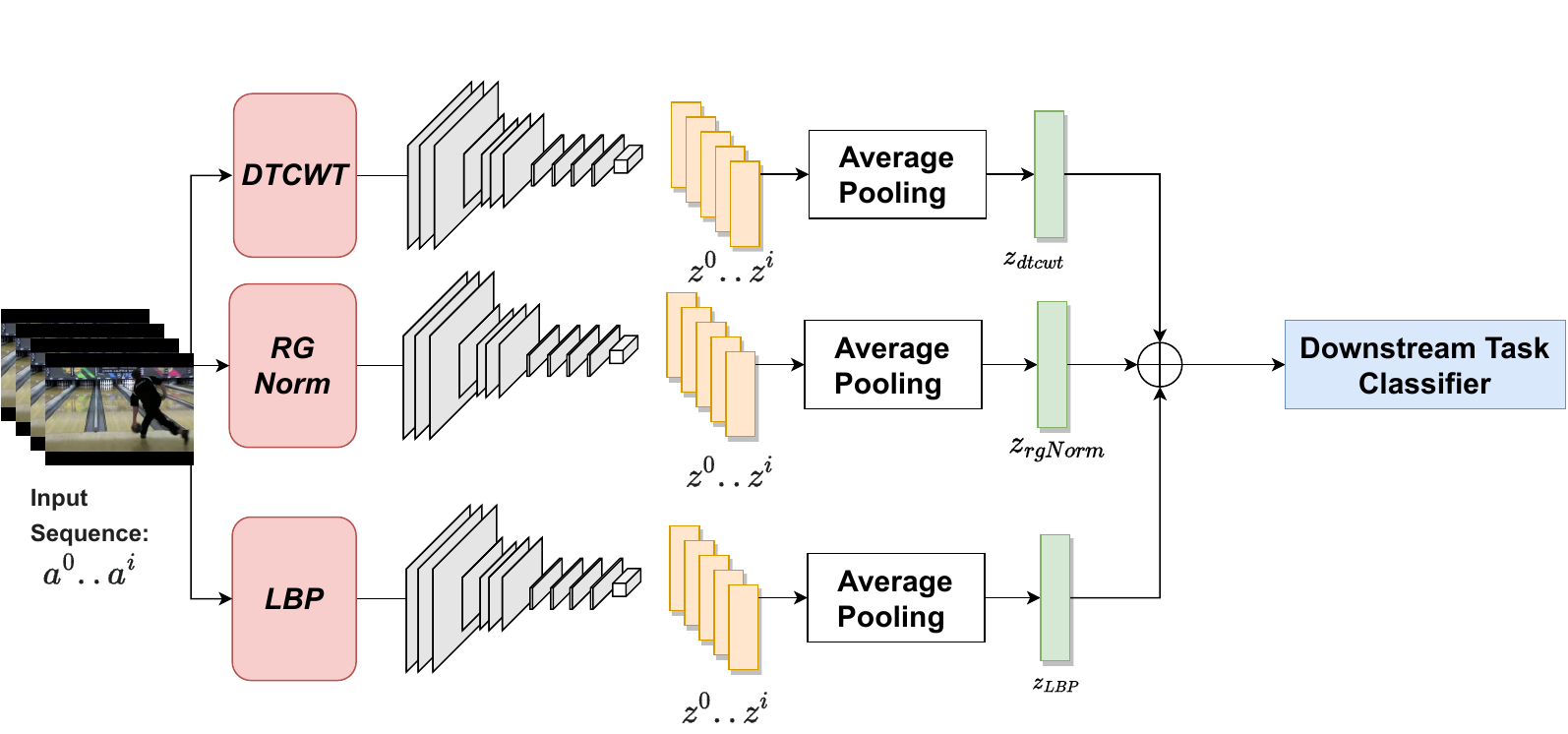}
	\caption{An illustration of the presented framework. Each encoder network is preceded by a transformation and is trained in a contrastive learning setting. Afterwards, the linear classifier is trained on a downstream classification task while the weights of the encoders are frozen.}
	\label{multimodal}
\end{figure*}

\subsection{Decomposition}
We explore the decomposition of the input signal using well-established operators: wavelet transform, rg normalization, and Local Binary Pattern (LBP) (Figure \ref{multimodal}). Each extracted representation is sensitive to different image features, creating different invariant spaces. These operators act as inductive biases to learn invariant latent representations. We hypothesize that our approach will leverage quasi-invariant spaces to learn more robust, stable, and generalizable representations, especially in the context of local plasticity rules where the signal cannot be backpropagated across the entire model. An extended version of our pipeline for video data, adding a motion-sensitive encoder, is in the appendix.

\textbf{LBP:} Local Binary Pattern (LBP) is a visual descriptor that captures local texture patterns in an image by comparing each pixel's intensity with its neighbors and assigning a binary number based on the comparison \cite{ojala2002multiresolution}. LBP is popular in applications like face recognition and texture analysis due to its robustness to illumination changes and computational simplicity \cite{maenpaa2004local, ahonen2006face}. To address rotation sensitivity, \cite{maenpaa2004local} suggests circularly rotating the neighboring pixels until the minimum binary value is obtained, ensuring consistent LBP values regardless of image rotation. We employ this rotation-invariant version of LBP in our experiments.


\textbf{RG Normalization:} To obtain color values without  intensity information, it is possible to transform RGB channels of an image to a normalized rg space \cite{GEVERS1999453}. This is helpful to reduce the effect of varying illumination levels onto the task. The normalized rg values can be computed in the following way:
\begin{align}
    r_{norm} = \frac{R}{R+G+B} \qquad g_{norm} = \frac{G}{R+G+B}
\end{align}

\textbf{Wavelet Transform:} The third decomposition operator we consider is the Dual Tree Complex Wavelet Transform (DTCWT), an enhancement of the discrete wavelet transform (DWT) that offers better shift invariance \cite{kingsbury1998dual}. Unlike Fourier transforms, which capture only frequency, DWT captures both frequency and location in time. DTCWT is similar to the Morlet transform used in invariant scattering convolution networks \cite{bruna2013invariant}. For detailed theory on different wavelet transforms and their advantages, see \cite{valens1999really}.

 
\subsection{Local Learning with Contrastive Predictive Coding}

We use Contrastive Predictive Coding (CPC) to train encoders before concatenating their representations for downstream tasks, as detailed in \cite{oord2018representation}. CPC involves the encoder learning to predict future responses (e.g., image patches or video frames) while distinguishing these from negative examples in latent space.

A follow-up work \cite{illing2021local} interprets this future response as eye movement fixations, suggesting that consecutive frames should have similar latent representations when fixating on the same object/scene. In this biologically inspired setting, the Contrastive, Local, And Predictive Plasticity (CLAPP) model uses a Hinge loss to train encoders layer-wise through a Hebbian-like learning rule. We use this CLAPP model for our experiments and compare it with a CPC variant trained end-to-end with Hinge loss, referred to as HingeCPC.

For image data, self-supervision sequences are created by dividing images into overlapping patches, simulating a sequence of movements. The encoder produces a representation \(z^t\) for each patch. These patches serve as context for others, determining if they stem from a continuous scene. The context \(c^t\) at time \(t\) predicts future representations \(z^{t+\delta}\) up to 5 patches ahead. Positive examples come from the same image or video, while negative examples are random patches from other images in the batch. The prediction uses a linear transformation matrix \(W^{pred}\):

\begin{align}
z_{pred}^{t}=W^{pred} \cdot c^t    
\end{align}

Predictions \(z_{pred}^{t}\) are aligned with actual representations \(z^{t+\delta}\) for positive examples or contrasted for negative ones using Hinge loss:

\begin{align}
    L^{t} = max(0,1- y^{t} \cdot z^{t+\delta} \cdot z_{pred}^{t} )  
\end{align}

Here, \(y^t\) is a binary classification label with \(y^t=1\) for positive and \(y^t=-1\) for negative examples.

Hinge loss can be applied to the last layer with backpropagation (HingeCPC) or computed layer-wise (CLAPP) for local Hebbian learning.


\section{Experiments}
\subsection{Experimental Setting}
\textbf{Training Details and Hyperparameters:} 
The encoders in our pipeline are 6-layer CNNs trained in parallel with two versions: \textbf{ours(Local)} using local learning rules, and \textbf{ours(Backprop)} using backpropagation.

We compare against three baselines: Supervised-VGG (end-to-end supervised classification), CLAPP (local plasticity rules for CPC), and HingeCPC (backpropagation with Hinge loss for CPC) \cite{illing2021local}.

Encoders are trained for 200 epochs using Adam with an initial learning rate of $10^{-5}$ and weight decay of $5 \cdot 10^{-6}$. We use a mini-batch size of 64, randomly cropping and downsizing images to 64x64, and applying random flipping with probability $p=0.5$. Training lasts for 200 epochs. For image datsets, Static images are split into 16x16 patches to simulate a sequence of frames.

\textbf{Datasets:} We evaluate our models using two standard classification benchmarks (GTSRB \cite{stallkamp2011german} and STL10 \cite{coates2011analysis}), one multi-target classification dataset of concrete defects (CODEBRIM \cite{mundt2019meta}), and an action recognition video dataset (UCF101 \cite{soomro2012ucf101}). Results for the video dataset are in the appendix.

\begin{table*}[t]
  \centering
  \begin{tabular}{@{}lrrrrr@{}}
    \toprule
    &\multicolumn{2}{c}{GTSRB} &\multicolumn{2}{c}{STL10}\\
    Method & Accuracy  & Difference  & Accuracy & Difference \\ 
    \midrule
    Supervised VGG       &   {\boldmath $97.4 \pm 0.42$} &   {-}  & {\boldmath $75.47 \pm 0.03$} &   {-}    \\

		HingeCPC \cite{illing2021local,oord2018representation}       &   {$84.2 \pm 0.7$}   &   {-} &   {$69.18 \pm 0.07$}   &   {-}  \\
		CLAPP \cite{illing2021local}        &   {$79.61 \pm 0.81$}  &   {-}    &   {$66.61 \pm 0.28$}  &   {-}  \\
		\hdashline
		{$LBP$}+ HingeCPC        &   {\boldmath $94.8 \pm 0.46$}  &   {\boldmath $+10.6$} &   {$53.52 \pm 0.12$}  &   {$-15.66$}   \\
		{$RGNorm$} + HingeCPC         &   {$35.78 \pm 0.16$}  &   {$-48.42$} &   {$46.24 \pm 0.49$}  &   {$-22.94$}  \\
		{$DTCWT$}+ HingeCPC         &   {$77.44 \pm 0.6$}    &   {$-6.76$}  &   {\boldmath $71.44 \pm 0.01$}    &   {\boldmath $+2.26$} \\
\hdashline
		{$LBP$} + CLAPP       &   {\boldmath $94.2 \pm 0.34$}  &   {\boldmath $+ 14.5$}    &   {$55.35 \pm 3.44$}  &   {$- 11.26$}    \\
		{$RGNorm$}+ CLAPP       &   {$33.78 \pm 0.45$}  &   {$-45.83$} &   {$45.59 \pm 1.45$}  &   {$-21.02$}     \\
		{$DTCWT$}+ CLAPP       &   {$76.1 \pm 0.51$}    &   {$-3.51$}  &   {\boldmath $69.37 \pm 0.01$}    &   {\boldmath $+2.76$}  \\
\hdashline
	\textbf{ours (Backprop)}	     &  \hlfancy{pink}{ $94.83 \pm 0.55$ }  &   {$+10.63 $}   &   \hlfancy{pink}{$74.13 \pm 0.05$} &   {$+4.95 $}   \\
	\textbf{ours (Local)}	      &   {$94.21 \pm 0.24$} &   \hlfancy{pink}{$+14.6$}    &   {$72.97 \pm 0.02$} &   \hlfancy{pink}{$+6.36$}   \\
    \bottomrule
  \end{tabular}
  \caption{Performance comparison in terms of classification accuracy (\%). The best performing self-supervised model is highlighted in red, in bold the best performing model for each category: baselines, models trained with HingeCPC, models trained with CLAPP, decomposed encoders. The difference column contains the performance gap between the model and its baseline, which is HingeCPC for for encoders trained with backpropagation and CLAPP for encoders trained with local plasticity rule.}
  \label{tableRes1}
\end{table*}

\subsection{Experimental Results}

\begin{figure*}[htbp]
	\centering
	\label{tsne1}\includegraphics[width=0.35\linewidth]{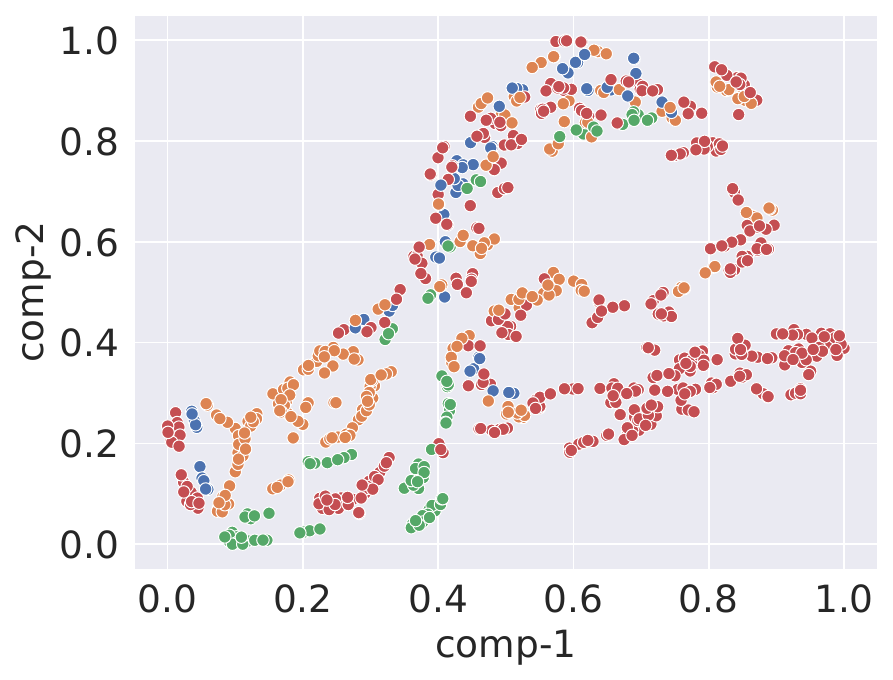}
	\label{tsne2}\includegraphics[width=0.35\linewidth]{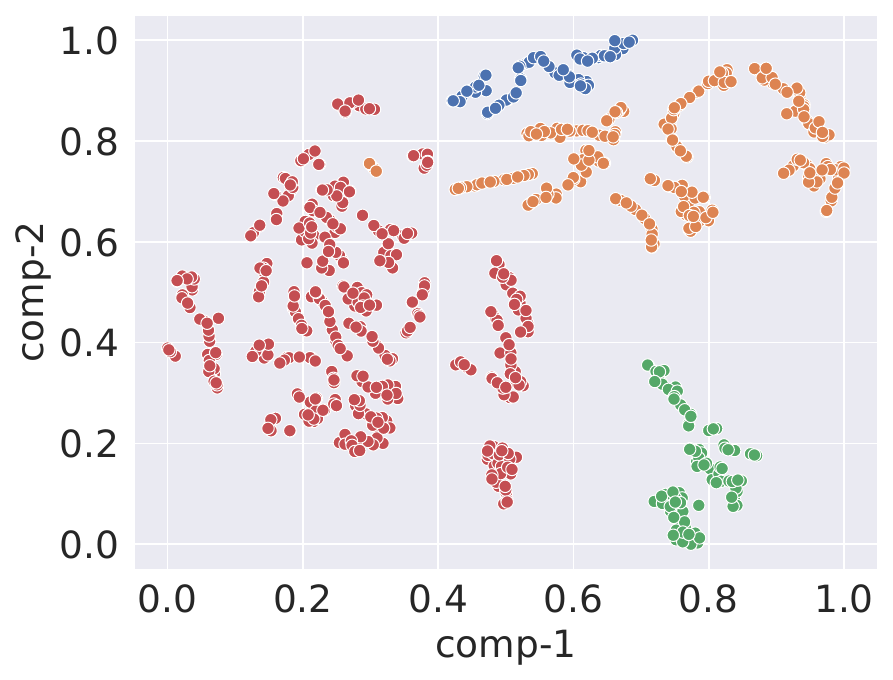}\qquad%

	\caption{Qualitative results illustrating t-SNE  dimensionality reduction of the latent encodings on GTSRB of  CLAPP (left plot) and  our framework trained locally (right plot) models. We visualize only 4 classes of street signs to avoid clutter. The following classes were randomly chosen: Speed Limit 20km/h sign (\hlfancy{soulblue}{blue}), Turn Straight Sign (\hlfancy{soulgreen}{green}), No way general sign (\hlfancy{soulorange}{orange} ), Attention bottleneck sign (\hlfancy{soulred}{red}). Further visualizations of t-SNE and PCA dimensionality reduction are included in the appendix.}
	\label{tsne}
\end{figure*}

The empirical results for classification task on STL10 and GTSRB are found in table \ref{tableRes1}. Both versions of our method  (Backprop and Local) achieve a performance closer to that of the supervised models. Ours(Local) sees a $14.6\%$ performance improvement on GTSRB and $6.36 \%$ on STL10 compared to CLAPP,  i.e a local learning variant without decomposition. Interestingly, the improvement due to decomposition is higher for local learning than for the layer-wise backpropagation: ours(Backprop) improves only by $10.63\%$ in comparison to HingeCPC. 

Hence, decomposition seems to help close the gap between learning with local plasticity rules and learning with backpropagation. Decomposition can be argued to serve as a powerful inductive bias to learn useful representations, even though error signals cannot be backpropagated across the whole network. Further, the visual  operator that contributes most to the performance improvement depends on the properties of the dataset: LBP for GTSRB and DTCWT for STL10.

We qualitatively support our empirical classification results (Table \ref{tableRes1}) by plotting the t-SNE \cite{van2008visualizing}  embedding of the encodings on CLAPP and on our method with decomposition (Figure \ref{tsne}). The plot shows the potential of our decomposition approach in separating representations of different classes. However, we note that separability of t-SNE clusters depends on the chosen dataset and the visualized classes. For instance, the class with speed limit sign 20km/h will form a cluster that is much closer to classes of other speed limit signs. 
\subsection{Ablations}
We conduct an ablation study to investigate the impact of concatenating the representations of multiple parallel encoders. We train three encoders with CLAPP loss in a similar manner to previous sections and then concatenate their latent representations for prediction.
The object classification results are presented in Table \ref{abl}. Indeed, representation concatenation seems to improve the performance compared to a single encoder, but the performance improvement is not as pronounced as the improvement with the additional operators (2\% compared to 6.48\% on STL10 and 4\% compared to 14\% on GTSRB). This further highlights that the improved performance of our proposed network is not primarily due to just multiple encoders but mostly due to strong inductive biases provided by the visual invariant operators. 

\subsection{Multi-Target Classification}

To further understand how of the decomposition approach scales to real world applications, we benchmark our method on the CODEBRIM dataset \cite{mundt2019meta}.  We train our encoders in a similar manner to previous experiments.  After that, we freeze the weights of the encoder and train a linear downstream multi-target classifier on representations created
by the frozen encoder (Table \ref{codebrimRes}).

We see that our method leads to a better classification performance on all CODEBRIM classes compared to standard end-to-end models. Our design achieves a multi-target performance that is close to supervised baseline  both with local rules and backpropagation. And this is despite training the encoder in a self-supervised setting. 

\begin{table}
  \centering
  \begin{tabular}{@{}lcr@{}}
    \toprule
    Method & $STL10$ & $GTSRB$\\ 
    \midrule
    CLAPP \cite{illing2021local}      &   {$66.61 \pm 0.28$}  &   {$79.61 \pm 0.81$}    \\

		Multi-enocder CLAPP     &   {$68.65 \pm 0.44$}  &    {$83.81 \pm 0.78$}   \\

		   \textbf{ours (Local)}   &   {$72.97 \pm 0.02$} &    {$94.21 \pm 0.24$}     \\

		\bottomrule

  \end{tabular}
  \caption{Ablation of the visual invariant operators. Performance comparison in terms of classification accuracy (\%)}
  \label{abl}
\end{table}

\begin{table}
  \centering
  \begin{tabular}{@{}lrr@{}}
    \toprule
    Method & Multi-Target & AvgAcc\\ 
    \midrule
    Supervised VGG        &   {$58.9 \pm 0.3$} &   {$88.87 \pm 0.46$}     \\
	    MetaQNN-1 \cite{mundt2019meta}   &   {\boldmath $66.2 \pm 1.6$} &   {\boldmath  $87.6 \pm 0.5$}     \\
		   HingeCPC \cite{illing2021local,oord2018representation}  &   {$33.87 \pm 0.12$}   &   {$82.27 \pm 0.27$}  \\
		CLAPP \cite{illing2021local}      &   {$37.78 \pm 0.5$}  &   {$83.05 \pm 0.6$}    \\
\hdashline
		{$LBP$} + HingeCPC      &   {$23.3 \pm 0.18$}   &   {$77.34 \pm 0.3$}      \\
		{$RGNorm$}  + HingeCPC     &   {$6.25 \pm 0.1$}   &   {$76.2 \pm 0.1$}     \\
		{$DTCWT$}  + HingeCPC      &   {\boldmath $47.2 \pm 0.2$}   &   {\boldmath  $84.7 \pm 0.4$}     \\
	\hdashline
		{$LBP$} + CLAPP      &   {$20.11 \pm 1.5$}   &   {$76.94 \pm 0.83$}      \\
		{$RGNorm$} + CLAPP      &   {$6.25 \pm 0.1$}   &   {$76.22 \pm 0.1$}     \\
		{$DTCWT$} + CLAPP      &   {\boldmath $48.9 \pm 0.7$}   &   {\boldmath $85.5 \pm 0.3$}     \\
	\hdashline

		\textbf{ours (Backprop)}      &   {$53.19 \pm 0.5$} &   {$87.55 \pm 0.25$}     \\
		\textbf{ours (Local)}      &  \hlfancy{pink} {$54.15 \pm 0.4$} &   \hlfancy{pink}{$87.79 \pm 0.4$}     \\

		\bottomrule
  \end{tabular}
  
  \caption{Performance comparison in terms of multi-target classification and class average accuracy (\%) on CODEBRIM test set. Multi-target accuracy refers to classification of all classes correctly in the image.  The best performing self-supervised model is highlighted in red, in bold the best performing model for each category: baselines, models trained with HingeCPC, models trained with CLAPP, decomposed encoders.}
  \label{codebrimRes}
\end{table}

Moreover, we observe that the pipeline with the local learning slightly outperforms the pipeline trained with backpropagation. One possible explanation is that the CODEBRIM dataset is a much smaller dataset than STL10 or GTSRB. As observed in previous works \cite{lagani2021hebbian}, training with Hebbian rules can in some cases be more sample efficient. Further results in the appendix show that our decomposed achieves better performance on adversarial and noisy examples.

\section{Conclusion}
In this manuscript, we presented a framework to study the impact of decomposition on representation learning with local plasticity rules and backpropagation. Our experiments indicate that decomposing input signals with well-understood image transformation operators can enhance the generalizability of latent representations and narrow the gap between local learning rules and backpropagation.

These results underscore the importance of domain knowledge and application context in choosing the decomposition pipeline, as different datasets require different quasi-invariances. Formalizing the choice of operators is a crucial next step for stable performance across domains. Complementary operators are essential to project input signals into complementary quasi-invariant spaces, improving performance in various contexts.
\section{Acknowledgments}
This work was supported by the German Federal Ministry of Education and Research (BMBF) funded projects 01IS19062 ”AISEL” and 16DHBKI019 "ALI".

%
%
\bibliographystyle{splncs04}
\bibliography{main}
\end{document}